\documentclass{article}

\usepackage[nonatbib, preprint]{neurips_2020}

\usepackage[utf8]{inputenc} 
\usepackage[T1]{fontenc}    
\usepackage{hyperref}       
\usepackage{url}            
\usepackage{booktabs}       
\usepackage{amsfonts}       
\usepackage{nicefrac}       
\usepackage{microtype}      
\usepackage{amsfonts}
\usepackage{graphicx}
\usepackage{physics}
\usepackage{biblatex} 
\usepackage{caption}
\usepackage{subcaption}
\usepackage[dvipsnames]{xcolor}

\title{Improvements to Gradient Descent Methods for Quantum Tensor Network Machine Learning}

\author{
  Fergus Barratt\\
  TyTN\\
  \texttt{fbarratt@umass.edu} \\
  \And
  James Dborin \\
  TyTN\\
  \texttt{james.dborin.18@ucl.ac.uk} \\
  \AND
  Lewis Wright \\
  TyTN \\
  \texttt{lewis.wright@kcl.ac.uk}
}

\begin{document}

\maketitle

\begin{abstract}
  Tensor networks have demonstrated significant value for machine learning in a myriad of different applications.
  However, optimizing tensor networks using standard gradient descent has proven to be difficult in practice.
  Tensor networks suffer from initialization problems resulting in exploding or vanishing gradients and require extensive hyperparameter tuning.
  Efforts to overcome these problems usually depend on specific network architectures, or ad hoc prescriptions.
  In this paper we address the problems of initialization and hyperparameter tuning, making it possible to train tensor networks using established machine learning techniques.
  We introduce a `copy node' method that successfully initializes arbitrary tensor networks, in addition to a gradient based regularization technique for bond dimensions.
  We present numerical results that show that the combination of techniques presented here produces quantum-inspired tensor network models with far fewer parameters, while improving generalization performance.
\end{abstract}

\section{Introduction}
Quantum Tensor Network Machine Learning (QTNML) is a new and rapidly developing area of research.
Insights from many-body quantum physics have been combined with machine learning principles, producing novel machine learning models with heretofore unseen properties.
This family of models has been applied to a variety of different machine learning problems, such as image classification \cite{stoudenmire_schwab,lei_wang_peps, novikov_exponential_machines}, language modelling \cite{miller_language_modelling}, and anomaly detection \cite{anomaly_detection}.
 







Despite the success of QTNML, the best method for optimizing tensor networks remains unclear.
The Density Matrix Renormalization Group (DMRG) family of optimization algorithms is used throughout the literature.
These techniques involve freezing a majority of the tensors whilst updating a small subset  at a time.
In some formulations pairs of tensors are contracted together.
The new combined tensor is optimized and split using a Singular Value Decomposition (SVD).
This is the procedure used in \cite{stoudenmire_schwab}.
The most obvious alternative is performing gradient descent on all tensors at the same time, which we will call \emph{Direct Gradient Descent} (DGD).
DGD is used in \cite{anomaly_detection}, where it is noted that the tensor networks are initialized multiple times in order for training to occur.

There are two main advantages of DMRG optimization compared to DGD.
Firstly, this method seems more effective at optimizing large tensor networks with many nodes.
DGD suffers from exploding or vanishing gradients, which is further explored in the theory section below. There are theoretical reasons to believe that DMRG updates do not suffer from this problem.
For example, when the tensors fulfil certain isometric conditions, they will not suffer from exploding or vanishing gradients \cite{tangent_space_updates}.

Secondly, DMRG methods naturally allow for automatic bond dimension selection within the tensor network.
Only a single hyperparameter, the singular value cut off, needs to be tuned.
When using DGD the bond dimension of the tensor network must be set in advance.
This creates an extremely large hyperparameter search space, over all different bond dimension combinations.
But in practice it is common to set all bond dimensions to be the same and simply search through a single set of bond dimension values. 

There are problems with DMRG optimization algorithms however.
They are hard to implement on popular machine learning frameworks, like tensorflow\cite{tensorflow} and pytorch\cite{pytorch}.
Although possible, it is certainly not as straightforward as implementing DGD and could present a stumbling block for widespread adoption of tensor network methods.
Furthermore it is not clear how to combine tensors trained using DMRG, and tensors trained using DGD in the same training procedure.
This is not an issue for models developed in \cite{stoudenmire_schwab}, but in \cite{lei_wang_peps} it is clear there are benefits to be gained by combining QTNML methods with standard machine learning elements such as convolutional layers.
Finally, there is no generalisation of DMRG algorithms for more complex tensor network structures, such as MERA\cite{mera, stoudenmire_mera} or Tree Tensor Networks\cite{leiwang_ttn}, which suit certain data structures better.

\subsection{Our Contribution}
In this paper we improve on the direct gradient descent scheme for tensor networks, and in doing so surpass the performance found in DMRG-inspired optimization algorithms.
We present:
\begin{itemize}
    \item[A] An initialization scheme for quantum tensor network models which prevents exploding or vanishing gradients in the training process. This is supported with both theoretical and empirical justification. Models trained with this initialization show an increase in generalization performance over those trained using DMRG-inspired methods \cite{stoudenmire_schwab}.
    \item[B] A scheme to regularize the bond dimension during the gradient update process. The bond dimension is allowed to grow to increase model complexity if needed, whilst also penalizing growth in the size of the model. Models trained with this scheme show a significant decrease in parameter count for no significant loss of accuracy. 
\end{itemize}

\subsection{Tensor Networks}

    Many QTNML methods follow a similar pattern.
    A feature map, $\phi$, is used to embed the input data vector, $\vec{x}$ into a much larger feature space,
    
    \begin{equation}
        \phi(\vec{x})_{\sigma_{1}\sigma_{2}\dots\sigma_{N}} = \phi(x_{0})_{\sigma_{0}} \otimes \phi(x_{1})_{\sigma_{1}} \otimes \dots \otimes \phi(x_{N})_{\sigma_{N}}.
    \end{equation}
    
    A linear transformation, $\mathbf{W}_{\sigma_{1}\sigma_{2}\dots\sigma_{N}}^{\delta}$, can be used to create a decision function, f,
    
    \begin{equation}
        \mathbf{W}\phi(\vec{x}) = f(\vec{x}),
    \end{equation}
    
    where $\delta$ is the dimension of the output space.
    The size of the transformation matrix, $\mathbf{W}$, is exponential in the size of the input vector, hence this transformation is not possible to perform in general.
    Instead the matrix $\mathbf{W}$ can be factorized into smaller tensors contracted together to approximate the larger tensor.
    This set of tensors and the contraction performed is known as a \emph{tensor network}.
    An example representation of $\mathbf{W}$ factorizes the original tensor as a product of matrices, $A_{i,\sigma_{i}}$, aptly called \emph{matrix product states},
    
    \begin{equation}
        \mathbf{W} = \mathbf{A}_{0, \sigma_{0}}^{\delta} \mathbf{A}_{1, \sigma_{1}}\dots \mathbf{A}_{N, \sigma_{N}},
    \end{equation}
    
    where the output leg is placed on the first matrix.

    More generally, tensor networks consist of a number of \emph{tensors}, connected by \emph{tensor contractions}. 
    For the purposes of this paper, tensors are multidimensional arrays of numbers. 
    A tensor network, when contracted fully, produces a new tensor with some number of indices. 

    A generic tensor network consists of $V$ tensors, $A_{\vec{v}_i}$, each indexed by a vector $\vec{v}_i$ of integers, taking on values in the space $\bigotimes_{j \in \mathbb{Z}^{r_i}} \mathbb{Z}^{D_i}$.
    The rank of tensor $A_{\vec{v}_i}$ can be written as $r_i = \mathrm{dim}(\vec{v}_i)$
    An arbitrary tensor network decomposition of a tensor $W_{\vec{\xi}}$, has the following form,
    \begin{equation}
        W_{\vec{\xi}} = \sum_{\vec{\sigma}} \prod_i^{V} A_{\vec{v}_i},\label{eq:eq1}
    \end{equation}
    where $\vec{\sigma} = \bigcup_{i}\vec{v}_i \setminus \vec{\xi}$.
    Such a tensor network has $E$ edges, with those edges connecting tensors $k$ and $l$ lying in the pairwise intersection $\vec{v}_k \cap \vec{v}_l$. 

\section{Initialization for Tensorization}
    One major application of tensor networks techniques is in \emph{tensorization} of deep learning models \cite{tensorizing_nns_novikov}. 
    Modern deep learning involves the use of deep learning networks with up to billions of parameters.
    Such liberal use of parameters makes it difficult to use such networks for inference in resource constrained environments: on mobile phones, microprocessors or practically sized cloud compute nodes.
    Many methods exist for reducing deep learning model size.
    Within the scope of \emph{model compression} one can use \emph{quantization}, \emph{pruning} and (or) \emph{clustering}, \cite{pruning_quantization} .
    Another field of techniques uses \emph{knowledge distillation} to convey knowledge from a large network to a small one, with an architecture that might be unlike the original \cite{knowledge_distillation}.

    Another complementary line of research, \emph{tensorization} is a method by which expensive dense or convolutional layers in deep networks are replaced by tensor network factorizations of the form Eq.~\ref{eq:eq1}.
    These replacements can lead to large reduction in parameter count, which can lead to speedups for inference. 

    One important question is the best way to train these tensorized layers. 
    One important tool in doing so is a means to initialize them effectively. 
    Initialization of linear layers in neural networks is a well studied topic.
    Most commonly used, and available in most modern software packages \cite{tensorflow,pytorch} are the \emph{glorot}\cite{glorot}, \emph{he}\cite{he} and \emph{lecun}\cite{lecun} initialization methods.
    In this section we will show how such methods can also be used for tensorized neural network initialization, with appropriate modifications.

    \subsection{Probabilistic Initialization for Generic Tensor Network Models}\label{sec:gen_tensor_networks}
    In a deep learning model, the initial elements of the matrices (whatever structure they be a part of) are usually drawn from some probability distribution.
    If the model's matrices are represented by tensor networks, we can no longer set the values of these matrices directly, as they only appear as the result of the full contraction of the tensor network. 
    We can only set the elements of the constituent tensors. 
    However, we will show that we can set the elements of $W$ probabilistically through cunning choices of the element distribution of these tensors. 

    \subsubsection{Theory}
        Tensor networks for machine learning are most often used to represent matrices, and it is this application that we will consider going forwards. 
        We will specialise further to the case where the \emph{bond dimension} $D_i$ of each index is the same $D_i = D~\forall~i$. 
        The generalisation to arbitary tensor decompositions and bond dimension distributions is natural. 

        In a tensorized deep learning layer, the weight matrix $W$ is represented by a sum of products of elements of the tensors $A$ making up the tensor network,
        \begin{equation}
            W_{ij} = \sum_{\vec{\sigma}} \prod_n A_{\vec{v}_n}.\label{eq:eq2}
        \end{equation}
        A chosen distribution for each tensor element will lead to another distribution when all tensors are contracted, i.e. elements of $W_{ij}$
        For simplicity, we will consider the case where each variable is drawn independent and identically from $P$.
        We will show that this assumption suffices to derive a simple and robust initialization scheme. 
        
        Let the elements of the tensors be drawn from a distribution $P$, such that $\mathbb{E}[A_{\vec{v}_n}] = 0$, and $\mathrm{Var}(A_{\vec{v}_n}) = \sigma^2$. What can we say about the distribution of $W_{ij}$?

        From Eq.~\ref{eq:eq2} and the independence of the variables $A_{\vec{v}_n}$, it is clear that $\mathbb{E}(W_{ij}) = 0$.

        Each term in the sum in Eq.~\ref{eq:eq2} is the product of $V$ variables drawn independently from $P$. 
        The variance of each term $X_i = \prod_n^V A_{\vec{v}_n}$ in the sum is therefore,
        \begin{equation}
            \mathrm{Var}\left(\prod_n A_{\vec{v}_n}\right) = \prod_n \sigma_n^2 = \sigma^{2V},
        \end{equation}
        where we have used the fact that the variables are i.i.d. 
        
        The elements $W_{ij}$ are equal to the sum of $D^E$ such random variables.
        The sum of $n$ uncorrelated variables has a variance equal to the sum of the variances of the constituent variables.
        For each pair $X_i, X_j$, there is at least one $n$ for which $\vec{v}_n^i$ and $\vec{v}_n^j$ differ. 
        As such, the different elements of the sum are uncorrelated, and the variance of their sum is,
        \begin{equation}
            \mathrm{Var}(W_{ij}) = D^E \sigma^{2V}.
        \end{equation}

        By choosing the mean and variance of the constituent distributions, we can make sure that the mean and variance of the resulting $W$ distribution takes on a reasonable value. 
        Matching the mean and distribution of common choices of initializer, for example \cite{he,lecun,glorot}, should produce an effective initialization scheme for tensor network decomposed layers.

    \subsection{Algorithm}\label{subsec:initialization for tensorizaion algorithm}
    To initialize a tensorized layer such that the distribution of the elements $W_{ij}$ is approximately normal, with mean $0$ and variance $\sigma_{\mathrm{target}}^2$, each $A_{\vec{v}_n}$ should be drawn i.i.d from $P$ such that,
    \begin{equation}
        \mathbb{E}(A_{\vec{v}_n}) = 0,
    \end{equation}
    and
    \begin{equation}
    \mathrm{Var}(A_{\vec{v}_n}) = \left(\frac{\sigma_{\mathrm{target}}^2}{D^E}\right)^{\frac{1}{V}}.
    \end{equation}
    By initializing the tensors with a distribution with these properties, the mean and variance of the distribution of the elements of $\mathbf{W}$ is chosen appropriately.
    Such a method suffices for $V$ and $E$ sufficiently small. 
    Tensorized layers naturally fall into this category, since the number of nodes scales as $O(\mathrm{log}(N_\mathrm{features}))$.
    Other uses of tensor networks in machine learning do not have this property however. In the next section we will discuss an appropriate generalization for this case. 

\section{Initialization for Quantum-inspired Models}\label{sec:initialization}
    Tensor networks have also been used to construct models whose architecture differs radically from standard deep learning setups. 
    Since such models take inspiration from tensor network methods as used in quantum physics, we will hereinafter term them \emph{quantum-inspired} models. 
    The simulation of quantum systems requires performing linear algebra in a space exponentially large in the number of subsystems -- for example, for 100 electrons, matrix-vector operations in a $O(2^{100})$ dimensional vector space would be required for a full accounting of the physics. 

    Tensor network methods allow for the simulation of an important subset of the physics of such systems. 
    These techniques have been used in a similar way in machine learning.
    Data is embedded into an exponentially large space, and the output of the model is a (massive dimensional) linear operation on the embedded data, represented as a tensor network.

    Initialization of quantum-inspired models presents new challenges. 
    The number of nodes in a tensor network used in this manner is on the order of the number of features.
    Even for the simplest datasets used in image classification, this can be on the order of $1000$. 
    With the conservative assumption that the number of edges ($E$) and nodes ($V$) grow linearly with the number of features and for bond dimension $D=2$.
    A quantum-inspired tensor network model classifying MNIST requires summing $O(\mathrm{max}(D_i)^E) = 2^{784}$ numbers, each consisting of the product of $V = 784$ numbers. 
    Naive initialization schemes for the elements of the tensors often lead to wild instability, since we must balance the product of an exponentially large and exponentially small number.

    As a result of these numerical problems, tensor network methods often use novel training methods, since SGD based training schemes for quantum-inspired techniques can be tricky to use, and practically can only be implemented when the number of features is sufficiently small.

    In the following, we detail a practical scheme for initializing such tensor networks that can sidestep the problems associated with this instability. 

\subsection{Generic Tensor Network Models}
    The probabilistic method of Sec.~\ref{sec:gen_tensor_networks} should still work in principle for quantum-inspired models. 
    What will work mathematically however might not work using floating point numbers with finite precision.
    To initialize a network of a practicable size might require setting the variance below the floating point epsilon, or beyond the largest floating point number. 

    We would like a method with the same theoretical guarantees, but where initialization is easier. 
    This can be achieved with the inclusion of \emph{stacked copy nodes}. 
    \begin{figure}
        \centering
        \includegraphics[width=0.45\linewidth]{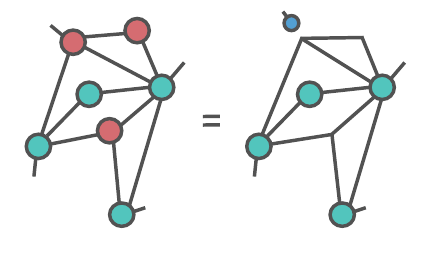}
        \caption{Copy node initialization. Replacing a fraction of the nodes of the tensor network with stacked copy nodes leads to an effective linear operator with known, manageable mean and variance. Red nodes are identified to be initialized as copy nodes. Red nodes with external legs are replaced by the tensor product of a copy node and a fixed vector.}\label{fig:copy}
    \end{figure}
    
    A \emph{copy node} is a multidimensional generalisation of the identity matrix. 
    It is defined by the following equation,
    \begin{equation}
        C_{ijk...} = \begin{cases}
                        1   \qquad i=j=k=...\\
                        0   \qquad \mathrm{otherwise}.
                    \end{cases}
    \end{equation} 
    A copy node in a tensor network has the effect of pinning all of the legs of that tensor to the same value. 
    One can naturally view a tensor network containing copy nodes as a \emph{hyper}-tensor network, where single edges can be connected to multiple nodes. 
    
    A natural means of handling the explosion of nodes and edges in quantum-inspired tensor networks is to initialize some number of the nodes in the tensor network by copy nodes.
    Then use the above prescription to initialize all of the other nodes, making sure that only a practicable number of non-copy nodes remain. 
    Replacing nodes in a tensor network with copy nodes results in a smaller hyper-tensor network. 
    The above discussion of random initialization of tensor networks is insensitive to the node (edge) structure of the network.
    The smaller embedded hypernetwork can be initialized by initializing its nodes according to the prescription of the respective tensor network.
    
    There is the question of definition for the input and output indices.
    To see the problem, consider the following simple example:
    If a node with both input and output indices is initialized with a copy node, the copy node will pin the input index on that node to the output node.
    The result is that the distribution of the output will not be as intended.
    Non-zero contributions to $W$ will only arise when the vector input to a single node is the same as the output of the network.
    
    The solution is to stack copy nodes across the input indices, and make sure nodes with output indices are densely initialized.
    We will show that this prescription leads to sensible initializations for the represented matrix. 
    
    \subsection{Algorithm}
    We will initialize an $n$-node subnetwork of our tensor network as described above, and all other indices with copy nodes.
    To do so, choose $V-n$ nodes at random from the tensor network, skipping any with an output edge\footnote{By assumption we can do always do this, since the output dimension is much smaller than the input}.
    Each node is initialized with a copy node in the following scheme.
    Replace all completely internal nodes with pure copy tensors. 
    For nodes with an input edge, initialize,
    \begin{equation}
        A_{\vec{v}_n}^{\xi_i} = C_{\vec{v}_n} \otimes \vec{v}_{\xi_i},
    \end{equation}
    where the internal indices are $\vec{v}_n$ as before, $\xi_i$ are input indices and $\vec{v}$ is a constant arbitrary vector (see Fig.~\ref{fig:copy}, where $\vec{v}$ is represented by the blue tensor).
    
    The result is an embedded initialization, where some subgraph of the tensor network is initialized.
    Some input nodes are initialized to fixed local vectors (in quantum terms, they are left in a product state), while others form part of the initialized subgraph.  This prescription is shown graphically in Figure \ref{fig:copy}. Red nodes are identified as being initialized with a copy node. Purely internal nodes are replaced with a copy node and nodes with an external leg are replaced with the tensor product of a copy node and a fixed vector. 
    
    The result is an initial value for the matrix $W$.
    Choose $v = (1, 0)$, and assume that the input tensors that are copy initialized are the first $k$ in some order\footnote{If they are not, the following matrix will have its rows permuted.}. 
    The resulting matrix will then have the following structure,
    \begin{equation}
        W_{tot} = \begin{pmatrix}
            W \\
            0 \\
        \end{pmatrix},
    \end{equation}
    where $W$ is the matrix of the densely initialized subgraph. 
    $W$ is initialized using the algorithm of Sec.~\ref{subsec:initialization for tensorizaion algorithm}, where $V$ and $E$ are the number of nodes and (hyper-)edges in the subnetwork. 
    
    Such a scheme is especially suitable for the feature map found in \cite{novikov_exponential_machines}.
    Since the initial network can be chosen to act only on the linear variables, so we can guarantee that the properties at initialization match those of the equivalent linear model.

\section{Automatic Bond Dimension Selection}

    In principle, for a tensor network with $E$ edges there are $E$ hyperparameters that should be set appropriately to define the model- a bond dimension $D_i$ for each edge. 
    In practise, given the difficulty of exploring such a large space of hyperparameters, all bond dimensions are set to a constant value and choices of the constant value are explored. 
    This strategy is used in \cite{lei_wang_peps} and \cite{stoudenmire_schwab}.
    
    This has several drawbacks. 
    A higher bond dimension between a pair of nodes leads to an ability to capture richer correlations between those tensors. 
    A high bond dimension might be required at several edges in a network to do this, while in other parts a smaller bond dimension might suffice. 
    If too small a global bond dimension is chosen, important correlations will be missed. 
    If too high a global bond dimension, it might well be able to capture important correlations at those places where it is needed.
    But the model will contain parameters at further edges that increase the model complexity but offer no additional explanatory power.
    
    These additional parameters can contribute to overfitting. 
    Furthermore, for applications on edge devices where either tensorization or quantum-inspired models offer important opportunities for parameter efficient machine learning, such unnecessary parameters increase power and compute requirements for no additional benefit. 
    
    Gradient based training for neural networks offers a natural solution to these problems.
    
\section{Rank Regularization}\label{sec: rank regularization}
    
    In this section we will detail a method for automatically choosing the bond dimensions of a tensor network model, which we term Rank Regularization (RR).
    
\subsection{Defining `soft' bond dimensions}
    In order to automatically determine an effective bond dimension for a network, we will relax the local bond dimensions of the network such that they can be trained as continuous parameters.
    These parameters can then be trained directly on the training data, or in the spirit of differentiable neural architecture search approaches \cite{DARTS}, we can define custom update steps for these parameters that depend on a held-out validation set.
    In this paper we will explore the first approach. 
    
\begin{figure}[h]
    \centering
    \includegraphics[width=0.44\linewidth]{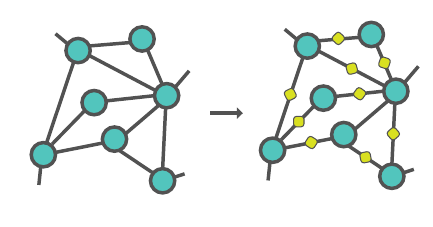}
    \includegraphics[width=0.26\linewidth]{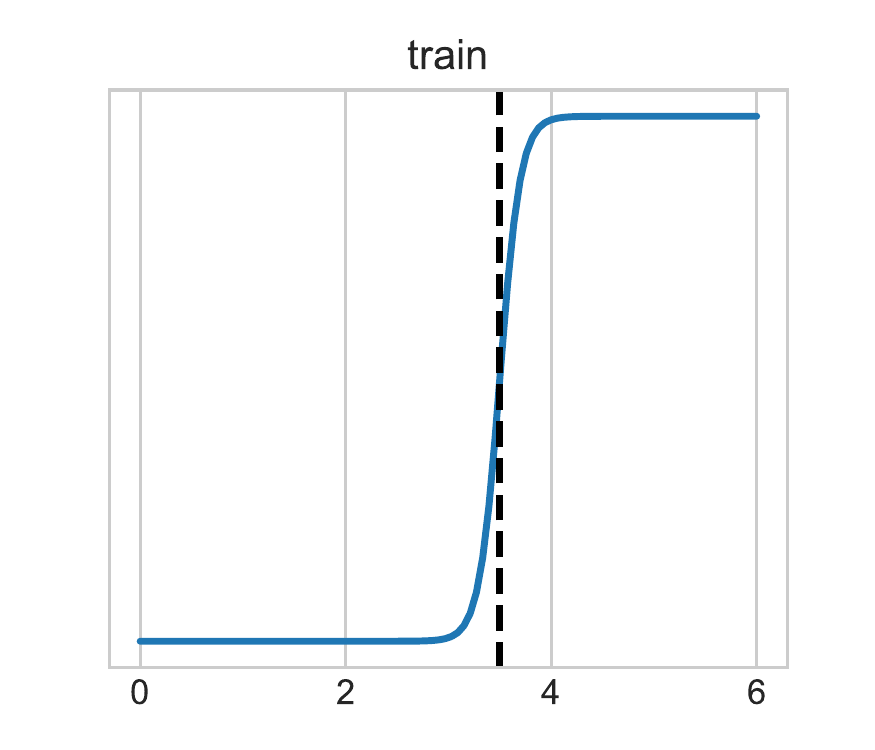}
    \includegraphics[width=0.26\linewidth]{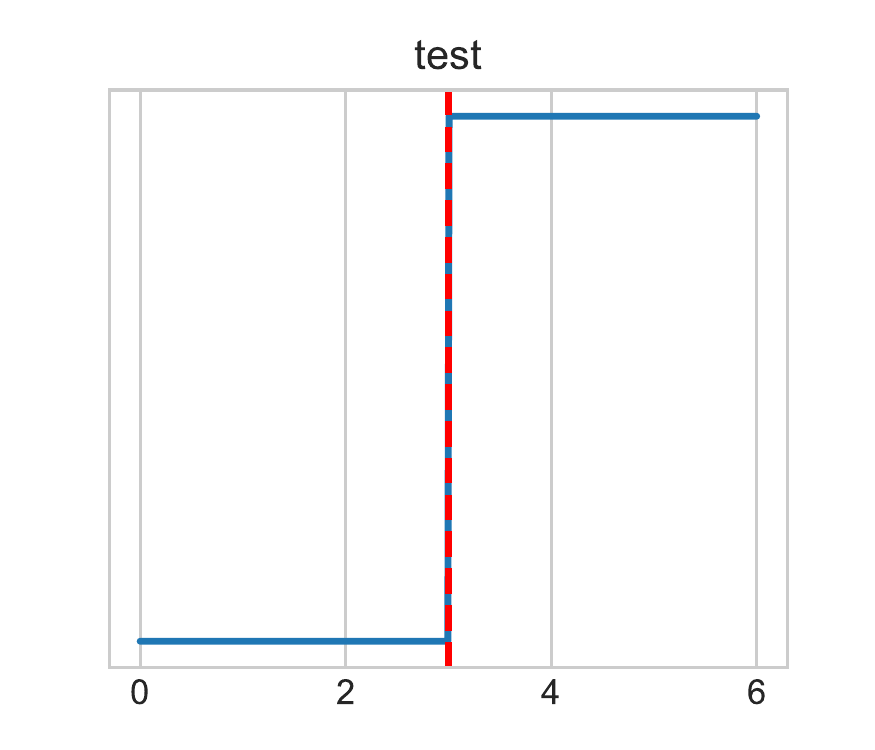}
    \caption{{Rank Regularizers.} a) A diagonal matrix, indicated by yellow diamonds, added to each internal edge in a tensor network, can be used to mask degrees of freedom in the tensor network, resulting in a reduction in the number of parameters. b) The form of the diagonal of the regularizers changes between training and inference, such that testing performance is that of the fully truncated network. The black dashed line indicates the soft bond dimension, whereas the red dashed line indicates the (rounded) final bond dimension choice.}
    \label{fig: rank reg}
    \end{figure}

    We term these relaxed bond dimension parameters 'soft' bond dimensions, and they are defined as follows. 
    At each edge on the tensor network model, define a 'rank regularizer', a diagonal matrix which is interposed between the tensors on that edge (see Fig~\ref{fig: rank reg}). 
    The original tensor network model can be recovered by setting each rank regularizer to the identity. 
    A family of truncated models can be defined by setting each rank regularizer to a projector with $D_i<D$ ones on the diagonal and $D-D_i$ zeros. 
    We can continuously explore this space of truncated models by using a shifted sigmoid function to define the diagonal of each rank regularizer, with a edge dependent 'soft' bond dimension defining the shift, see Fig.~\ref{fig: rank reg}.
    
    At training time, each soft bond dimension is variationally updated with the gradient of the loss function. 
    Smaller soft bond dimension on some edges of the network will nullify the gradients of certain parameters.
    But if unmasking these parameters will lead to an increase in training performance, the bond dimension can increase.
    Importantly, gradient based training allows us to bias the training of the model towards models with smaller bond dimensions on different sites. 
    
    This can be done by adding a penalty to the loss function, for example an L1/L2 regularization on the vector of soft bond dimensions. 
    The regularization strategy will define the final distribution of bond dimensions. 
    We have found one effective strategy to be to compute a softened approximation to the number of parameters in the tensor network, by computing that metric as if each bond dimension was equal to its softened value. 
    
    Importantly, every step described so far can be performed differentiably with tools currently provided in standard software libraries. 

    Before using the network for inference, the model can be truncated at each edge., choosing a bond dimension that removes the elements of the tensors that are masked by the regularizers.
    Further, the rank regularizers can be simply absorbed into neighbouring tensors, such that the Floating Point Operations Per Second (FLOPS) cost of the forward operation of the model is no greater than the model without the regularizers (in fact, if any truncation is performed, the cost in both number of parameters and FLOPS is strictly lower). 
\section{Numerical Results}
Ref.~\cite{stoudenmire_schwab} is arguably the most well known application of tensor networks found in Machine Learning.
Therefore, we consider the same experimental setup to benchmark our results.
\subsection{Experimental Setup}
We consider the image classification problem of the MNIST data set- $28\times28$ grey scaled images of digits from zero through to nine.
Each image is fed through a $\left(2,2\right)$ averaged pooling layer shrinking the image to $14\times14$ pixels, before being flattened using a zig-zag approach.
Each pixel value is embedded int feature space using a localised feature map of the form form: for the $j$th pixel,
\begin{equation}
    \phi(x_{j})_{\sigma_{j}} = \left[ \cos(\frac{\pi}{2}\hat{x}_j) + \sin(\frac{\pi}{2}\hat{x}_j) \right],
\end{equation}
where $\hat{x}_j$ is the normalized pixel value.
No bias in used in any of our results, and no further modifications were made to the dataset.
The feature data is inputted into a Matrix Product State Tensor Network (MPS) which is then trained using stochastic gradient descent.
The output leg of the MPS consists of un-normalized predictions with respect to each class, which are turned into class probabilities with a softmax function.
\\
\\
Tensor network operations are implemented by Quimb \cite{Quimb} and trained using TensorFlow Keras \cite{tensorflow} for 100 epochs, with a batch size of $32$ and a $80:20$ training validation split.
We choose to train our tensor networks using Adam \cite{kingma2014adam}, with $\alpha = 0.001$, to optimize a categorical cross entropy cost function.
Each tensor network is initialized and trained 10 times using the respective initialization scheme. 
We show the average result alongside error bars for 1 standard deviation, i.e. $\pm \sigma$.
\subsection{Initializers}

\begin{figure}[ht]\hspace{-2.5cm}
     \centering
     \begin{subfigure}[b]{0.45\textwidth}
         \centering
         \includegraphics[width=\textwidth]{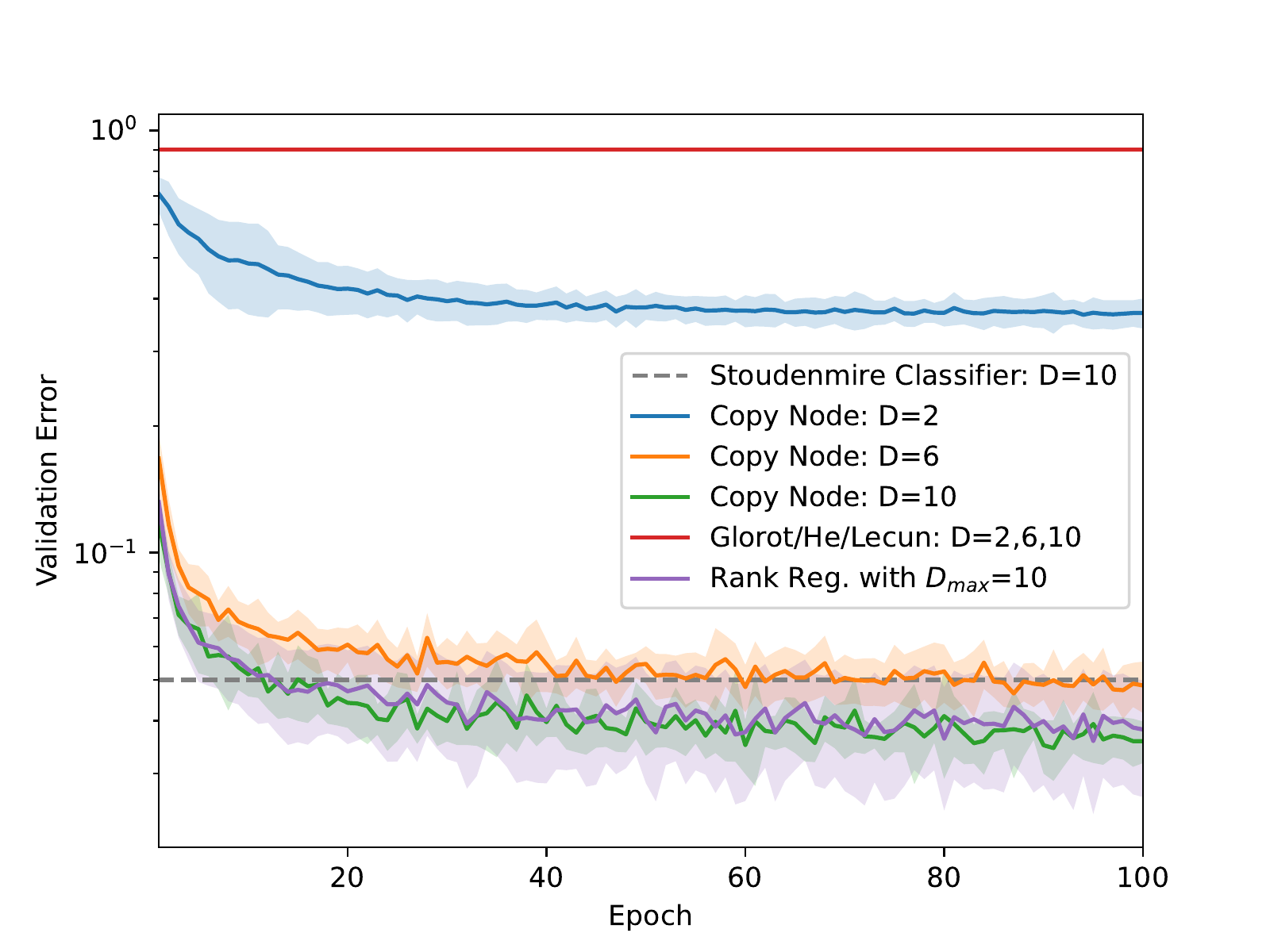}
         \caption{}
         \label{subfig:initializers}
     \end{subfigure}
     \begin{subfigure}[b]{0.45\textwidth}
         \centering
         \includegraphics[width=1.34\textwidth]{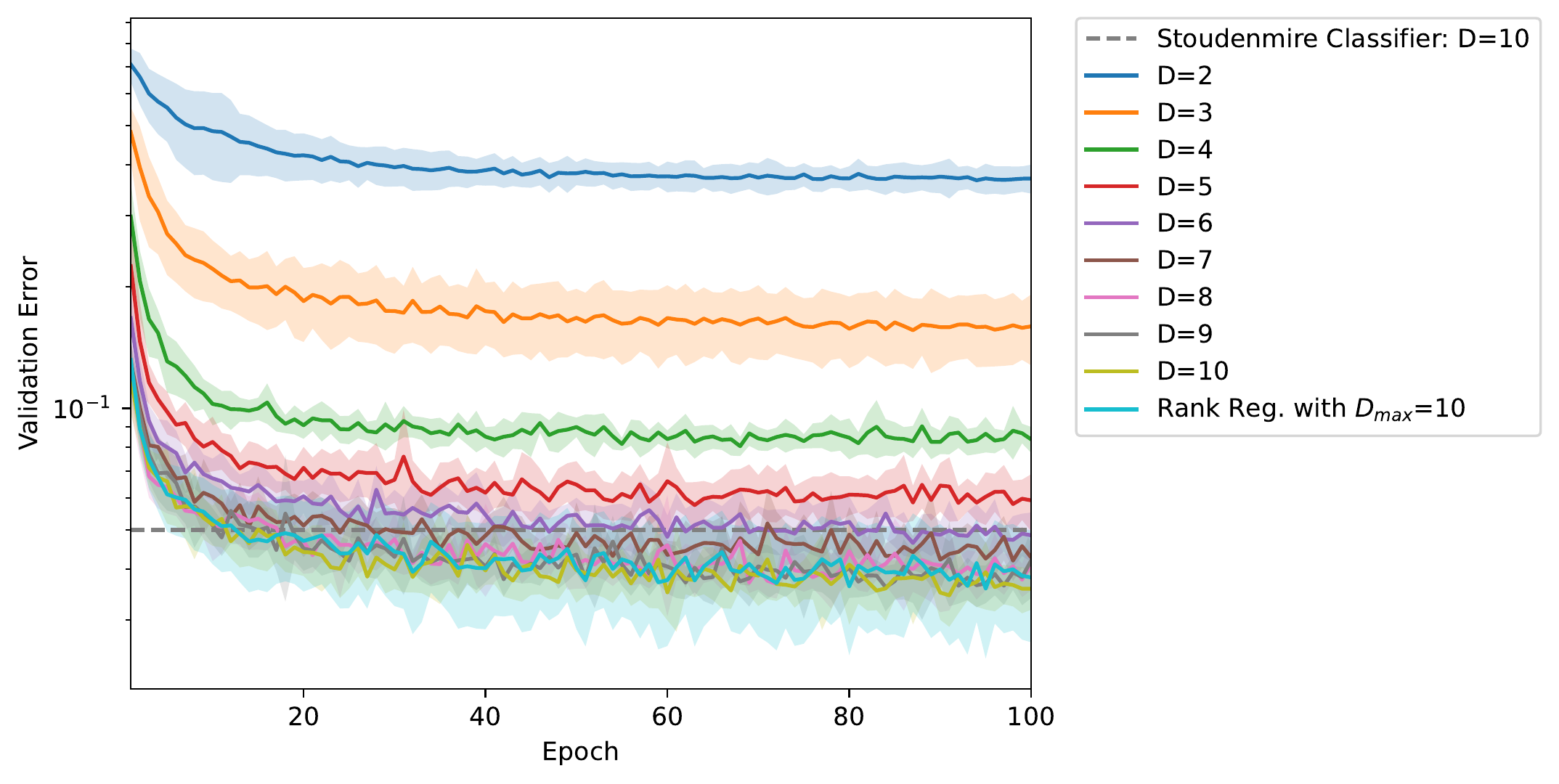}
         \caption{}
          \label{subfig:Bond Orders}
     \end{subfigure}
        \caption{Different MPS initializations results. a) Comparison of validation error during training between Glorot, He, Lecun and copy node initialization for $D = 2, 6, 10$. b) Validation error during training of copy node initialization for $D = 2, ..., 10$. Additionally, validation error for bond regularized MPS is also shown in both cases for $D_{max} = 10$. Our results outperform the different initialization schemes and \cite{stoudenmire_schwab} for $D \ge 6$.}
\end{figure}

Fig.~\ref{subfig:initializers} shows the average validation error during training of each MPS with different initializations and bond dimensions. 
Using either a Glorot, He or Lecun initialization of the MPS leads to a plateau in the loss and validation error for any bond dimension.
On the other hand, using the copy node initialization scheme discussed in section~\ref{sec:initialization} leads to a drop in validation error for all bond dimensions- as seen explicitly in Fig.~\ref{subfig:Bond Orders}.
We can see our initialization and training scheme start to outperform \cite{stoudenmire_schwab} in $\sim 20$ epochs for our equally sized $D=10$ and rank regularized classifier.
We observe no sign of over fitting due to the natural regulatory properties of Tensor Networks.

\subsection{Rank Regularization}
\begin{figure}[ht]
     \centering
     \begin{subfigure}[b]{0.45\textwidth}
         \centering
         \includegraphics[width=\textwidth]{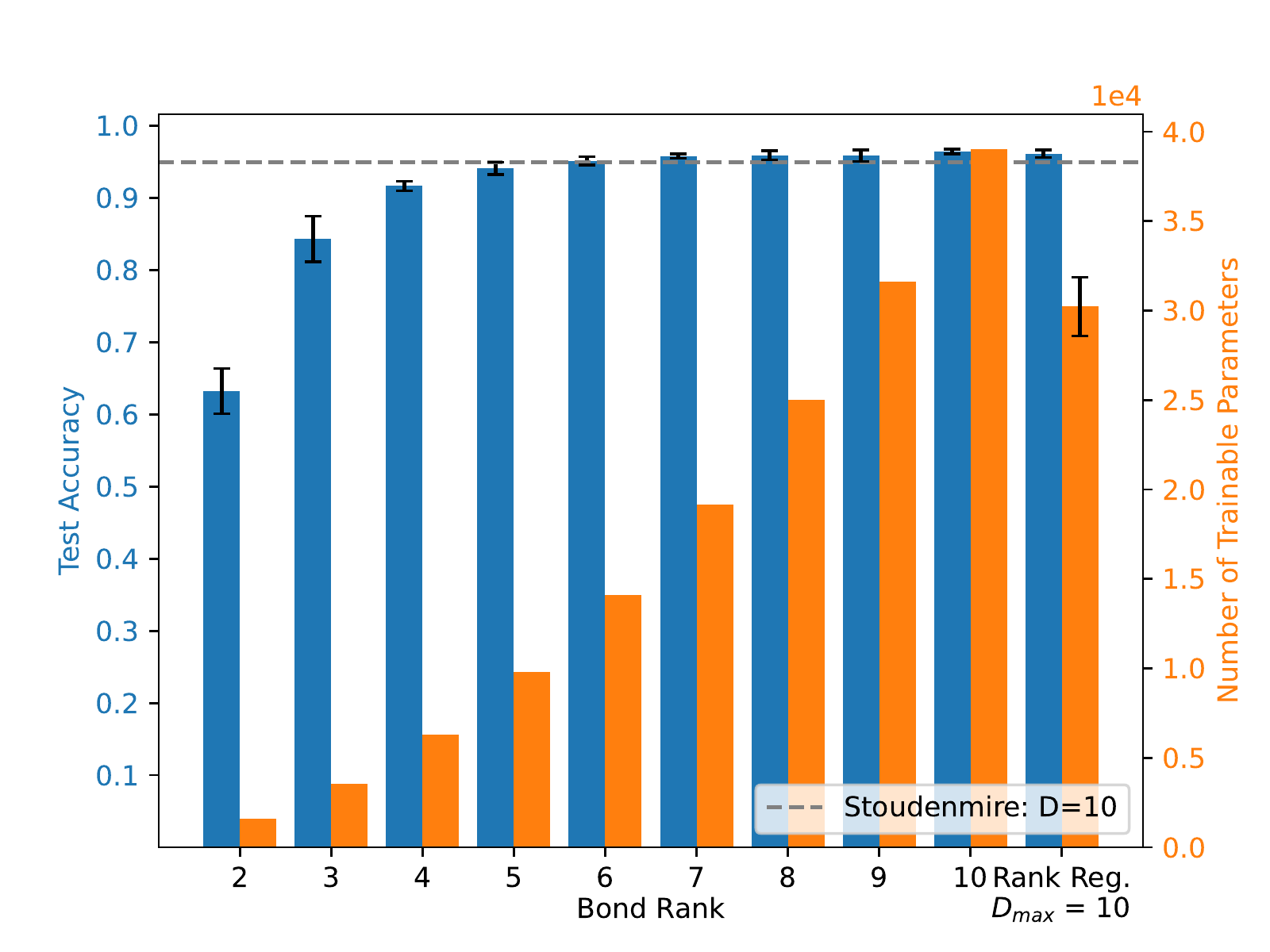}
         \caption{}
         \label{subfig:test accuracy and num params}
     \end{subfigure}
     \hfill
     \begin{subfigure}[b]{0.45\textwidth}
         \centering
         \includegraphics[width=\textwidth]{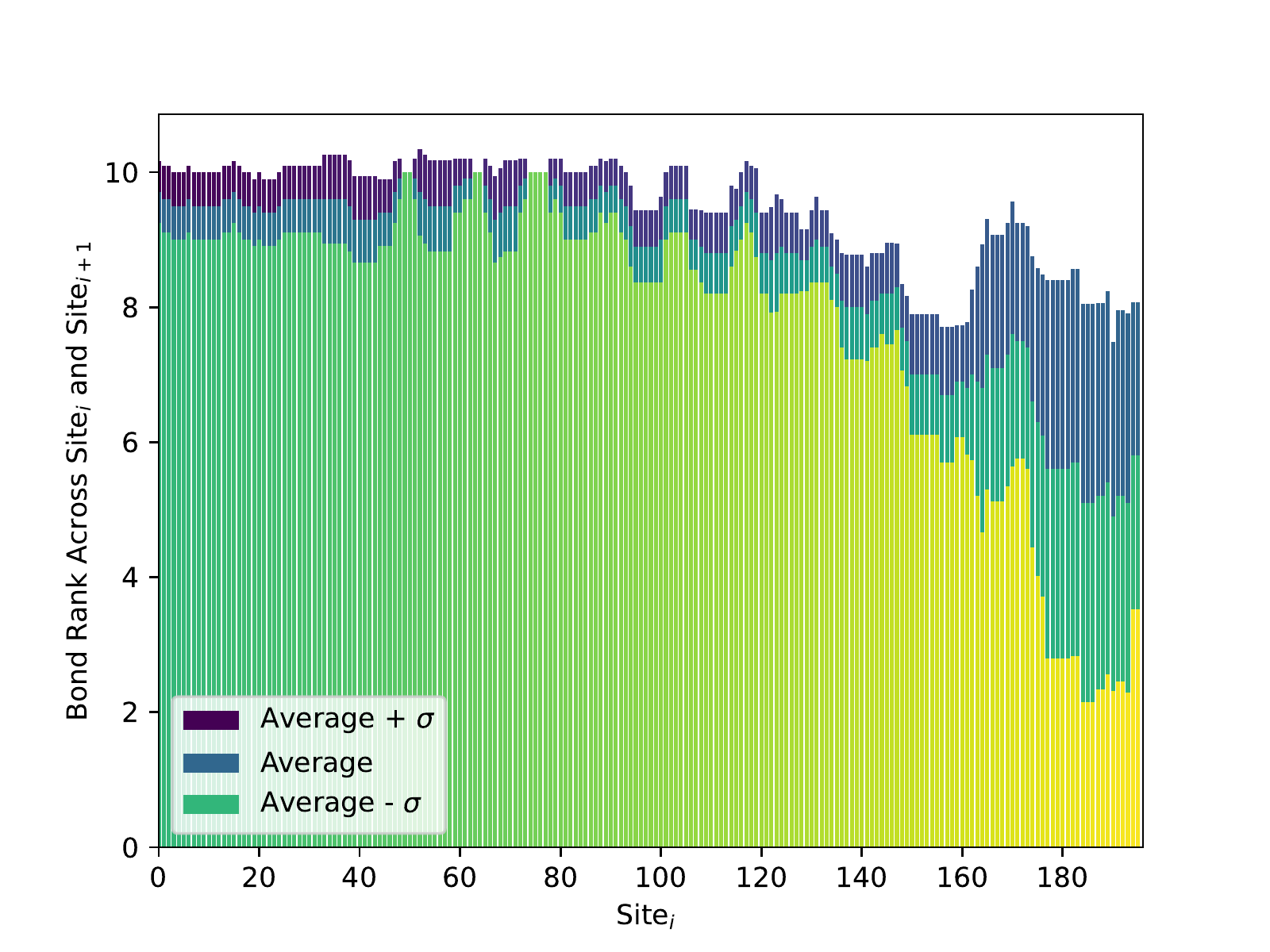}
         \caption{}
         \label{subfig:transformer spectrum}
     \end{subfigure}
        \caption{MPS model performance and bond regularization spectrum. a) Comparison of test accuracy and number of parameters for $D = 2, ..., 10$ MPS in addition to bond regularized MPS with $D_{max} = 10$. Our rank regularization scheme removes redundancies whilst performing equally as well as the $D = 10$ MPS. b) Spectrum of bond dimensions between neighbouring sites along the bond regularized MPS chain.}
\end{figure}
Fig.~\ref{subfig:test accuracy and num params} shows the test accuracy and number of training parameters for each MPS of different bond dimensions, initialized using the copy node scheme.
We see the expressive power of our classifier, outperforming \cite{stoudenmire_schwab} with a bond dimension of $D = 6$ and $\sim 64\%$ less parameters.
Additionally, we show that our bond regularization scheme, discussed in section~\ref{sec: rank  regularization}, for $D_{max} = 10$ performs equally as well as a $D = 10$ MPS with $\sim 22\%$ less parameters.
Further increases in bond dimension increase classifier performance further whilst still having less parameters than \cite{stoudenmire_schwab}.
\\
\\
The dimension of each regularized bond, $D_i$, can be seen in Fig.~\ref{subfig:transformer spectrum}.
The spectrum shows a drop in dimension towards the opposite end of the MPS chain with the output leg, since a lower dimension is needed to satisfy the amount of necessary information. 
We expect the peaks in the spectrum correspond to highly-valuable pixels shared by multiple digits.
Therefore a higher resolution i.e. bond dimension is needed to differentiate between them. 

\section{Conclusions}
In this paper we present two methods to improve the training of QTNML algorithms, which make it possible to train large tensor networks by performing gradient descent directly on all of the tensors.
We have introduced the \emph{copy node} initialization scheme, whereby certain tensors in the network are replaced by the tensor product  of a copy node and a constant vector.
The remaining nodes are then initialized such that when contracted they share the mean and variance with popular deep learning initialization distributions.
We demonstrate that large MPS are able to be trained with this method without relying on DMRG-inspired updates, and without exploding or vanishing gradients.
This method also extends to Tree Tensor Networks, MERA and PEPS.
We also introduce a method to dynamically increase or decrease the bond dimension during training by introducing masked diagonal tensors between each node in the tensor network.
Models trained with adaptive bond dimensions are able to achieve equivalent performance to higher bond dimension models with fewer parameters, focusing computational resources to the most important nodes in the classifying network.
With these enhancements standard gradient descent is now a viable alternative to DMRG-inspired training procedures. These improvements should make it easier to develop tensor network machine learning algorithms for a wider variety of tensor network geometries.

We have shown that with two modifications to stand SGD, quantum tensor networks can be effectively and efficiently trained to solve machine learning problems.
It would be interesting to explore the impact of these methods on a wider class of QTNML problems. 
It is well documented that the type of optimization algorithm that is used has a big impact on the quality of the trained model.
We give evidence that ordinary stochastic gradient descent on all tensors simultaneously can produce higher accuracy models than DMRG equivalents by demonstrating this on the MNIST dataset, which is a good indication of success on more complicated datasets.

\clearpage
\printbibliography

\end{document}